\pdfoutput=1

\documentclass[11pt]{article}

\usepackage{acl}
\usepackage{xcolor}

\usepackage{times}
\usepackage{latexsym}

\usepackage[T1]{fontenc}

\usepackage[utf8]{inputenc}

\usepackage{microtype}
\usepackage{todonotes}

\title{FERMAT: An Alternative to Accuracy for Numerical Reasoning}

\author{Jasivan Alex Sivakumar \and Nafise Sadat Moosavi \\
  Department of Computer Science \\ University of Sheffield \\ United Kingdom\\
  \texttt{\{jasivakumar1|n.s.moosavi\}@sheffield.ac.uk}}
  
\begin{document}
\maketitle
\begin{abstract}
While pre-trained language models achieve impressive performance on various NLP benchmarks, they still struggle with tasks that require numerical reasoning. Recent advances in improving numerical reasoning are mostly achieved using very large language models that contain billions of parameters and are not accessible to everyone. 
In addition, numerical reasoning is measured using a single score on existing datasets.
As a result, we do not have a clear understanding of the strengths and shortcomings of existing models on different numerical reasoning aspects and therefore, potential ways to improve them apart from scaling them up.
Inspired by CheckList \citep{ribeiro-etal-2020-checklist}, we introduce a multi-view evaluation set for numerical reasoning in English, called FERMAT.
Instead of reporting a single score on a whole dataset, FERMAT evaluates models on various key numerical reasoning aspects such as number understanding, mathematical operations, and training dependency. Apart from providing a comprehensive evaluation of models on different numerical reasoning aspects, FERMAT enables a systematic and automated generation of an arbitrarily large training or evaluation set for each aspect.
The datasets and codes are publicly available to generate further multi-view data for ulterior tasks and languages.\footnote{\url{https://github.com/jasivan/FERMAT}}\end{abstract}

\section{Introduction}
\noindent Numerical reasoning is an aspect that is often forgotten despite being an integral part of natural language. It is the ability to interact with numbers using the fundamental mathematical properties and thus model an area of human cognitive thinking \citep{saxton-etal-2018-analysing-NN-models}. Better understanding of numbers in language models would benefit various tasks like fact-checking \citep{vlachos-riedel-2015-identification}, text generation \citep{Moosavi-etal-2021-scigen, suadaa-etal-2021-towards}, and educational tools \citep{Mandal-etal-2022-ITS}. Current models' performance are still too weak with respect to numerical accuracy to then be used in downstream tasks like Infotabs \cite{gupta-etal-2020-infotabs} which requires identifying numbers in tables and then performing operations to correctly label statements causing factuality errors in such tasks. 

Recently, we have observed improved performances on relevant datasets about numerical reasoning using very large language models \cite{Wei-etal-2022-chain-of-thought, Lewkowycz-etal-2022-Minerva, kojima-etal-2022-zero-shot}. However, there are two main limitations to this recent trend. First, as models become larger their access becomes restricted to fewer users, i.e., users with the computational resources of large companies. For example, using one of the best mathematical models, the 540B parameter model Minerva \cite{Lewkowycz-etal-2022-Minerva}, would require over 2212G of memory for inference only. Second, the numerical reasoning capabilities of existing models are measured using a single score, i.e., mostly accuracy on common benchmarks like GSM8K \citep{Cobbe-etal-2022-gsm8k}. Therefore, their strengths and shortcomings in different aspects of numerical reasoning compared to other models are not clear. As a result, it is unclear what numerical reasoning aspects should be improved to improve their performance on datasets requiring numerical reasoning.

Motivated by CheckList \citep{ribeiro-etal-2020-checklist}, which is a behavioral test set concerning various linguistic aspects of the input language, we propose a unique and open \textbf{F}lexible \textbf{E}valuation set for \textbf{R}epresentating \textbf{M}ultiviews of \textbf{A}rithmetic \textbf{T}ypes,\footnote{We use the terms type, aspect and view interchangeably.} \textbf{FERMAT}, for evaluating the numerical reasoning capabilities of models based on multiple key aspects . 
It evaluates models according to (a) different ranges and representations of numbers, (b) different mathematical operations, and (c) the dependence of models on the fine-tuning data. In addition, it contains a tool to automatically generate new instances for each of its aspects. 
FERMAT enables (a) the identification of the strength and shortcomings of models according to its aspects, and (b) the automatic creation of additional training and evaluation instances using expert written templates that reflect FERMAT's categories.

FERMAT complements the recently proposed L\={I}LA benchmark \cite{Mashra-etal-2022-Lila} for mathematical reasoning.
L\={I}LA evaluates high-level aspects, e.g. whether performing mathematical reasoning also depends on commonsense knowledge or how the performance changes depending on the difficulty of the input language.
However, even the best-performing model on the L\={I}LA benchmark, i.e., a 2.7B parameter model that is fine-tuned on mathematical datasets, only achieves an accuracy of around 20-30 points when the input is formulated using a simple language and the test data is from a different distribution than that of the training, and it is not clear how to further improve this performance.

FERMAT, on the other hand, takes a deeper look at more fine-grained aspects by diving into the core mathematical abilities of the models and reporting which specific operations a model can or cannot perform and on which numbers. It also provides templates for creating more instances for each aspect, e.g., to generate additional data to further train or evaluate models on certain aspects.
FERMAT formulates the evaluation of numerical reasoning using the question answering format, which is commonly used in NLP for evaluating various skills \cite{Tafjord-etal-2019-Quarel, dasigi-etal-2019-quoref, jin-etal-2019-pubmedqa}.

We use FERMAT to highlight that single accuracy scores fail to give a holistic understanding of a model, that template diversity has a high impact in improving performance, and that number encodings play an important part in numerical reasoning.
The FERMAT framework could subsequently be adapted for different tasks according to the target application,\footnote{For instance, by automatically converting our QA templates to NLI \citep{demszky2018transforming} if NLI is a more suitable format for the downstream task.} to give a more targeted approach to improving models. Moreover, while the expert-written templates in FERMAT are written in English, they can easily be translated to be adapted to other languages.

\section{Related Work}

\subsection{Datasets}
Mathematical datasets focus on exploring different levels of difficulties and areas of maths. Some look at general symbolic maths, where the questions at least involve algebraic notations.
A certain group of datasets explores numerical reasoning in context, but the answers may not exclusively be numerical.
Unlike FERMAT, all these datasets evaluate models' performances on the whole dataset based on a single score.
Moreover, as a result of the availability of many datasets, new benchmarks have also been created based on regrouping the existing datasets according to specific criteria. Such benchmarks are created based on high-level aspects, e.g., how the performance changes when solving maths also depends on commonsense reasoning, when the maths is presented using equations, a simple language, or a complex language, or when the input is presented using a different task format. However, the performance of existing general-purpose models is very low, even on the simplest aspects, e.g., when the maths is presented using a simple language without requiring external knowledge. FERMAT, on the other hand, focuses on a fine-grained analysis of numerical reasoning by aiming to decipher models' ability to understand numbers, operations, and their reliance on the training data.

\subsubsection{General maths}
Dolphin18K \citep{huang-etal-2016-dolphin}, DeepMind Mathematics \citep{saxton-etal-2018-analysing-NN-models} and AQUA \citep{ling-etal-2017-aqua} are datasets that have a focus on solving algebraic problems and therefore use algebraic notation. 
These datasets are too complex for existing general purpose language models, mainly because they expect multi-hop reasoning.\footnote{E.g.
$[(6 \times 8)-(3 \times 6)] \div (6+4)$ \cite{ling-etal-2017-aqua}.}
For instance, \citet{Wei-etal-2022-chain-of-thought} only report an accuracy around 25\% for AQUA with a large, 62B parameter, model. 

\subsubsection{Numerical context}
Instead of the algebraic notation, some datasets are worded problems but are formulated as multiple choice questions, e.g. McTaco \citep{zhou-etal-2019-mctaco} and AQUA. This multiple choice format simplifies the task into a classification which prevents working with the continuous essence of numbers. Even if these are formatted into generative output tasks they then sometimes expect textual outputs like DROP \citep{dua-etal-2019-drop}. DROP has textual answers that can be extracted from the context which, similarly to the multiple choice questions, are disjoint from the numerical reasoning skill. 


\subsubsection{Numerical solutions}
The only datasets with textual input that solely expect numerical answers are GSM8K \citep{Cobbe-etal-2022-gsm8k}, MAWPS \citep{koncel-kedziorski-etal-2016-mawps}, CommonCore \citep{roy-roth-2015-CC} and Illinois \citep{roy-roth-2016-illinois}. GSM8K provides textual explanation for the solutions which has been effectively used by \citet{Wei-etal-2022-chain-of-thought}. However, similar to AQUA, GSM8K is very difficult for general purpose language models with reported results below 5\% accuracy using an 8B parameter model \cite{Wei-etal-2022-chain-of-thought}. Likewise, MAWPS requires some use of algebra to solve the problems. However, CommonCore and Illinois, which are subsets of MAWPS, are constituted of simpler one or two-hop problems.\footnote{An $n$-hop problem is one with the combination of, at most, $n$ of the basic operations.} Since FERMAT is designed to gain better insight by focusing on more accessible problems, CommonCore and Illinois are the ideal datasets. 

\subsubsection{View-based evaluation sets}
\citet{ribeiro-etal-2020-checklist} explain the motivation to move away from raw accuracy but towards more informative evaluation sets which give better insight into a given model. They look at different aspects of a test set; the skills needed to correctly solve the problem, in their case, linguistic phenomena like negation in sentiment analysis.

NumGLUE \cite{mishra-etal-2022-numglue}, on the other hand, is a multi-task benchmark that involves numerical reasoning. It combines different tasks like commonsense, domain specific language, quantitative expressions, with arithmetic understanding to create a more challenging benchmark. It also uses different question format such as fill-in-the-blanks, textual entailment, multiple choice questions, span extraction and numerical outputs.

A more mathematically expansive set is the recently introduced L\={I}LA dataset \citep{Mashra-etal-2022-Lila} where they regroup 20 existing datasets into 23 reasoning tasks including some of NumGLUE. These tasks are split into maths domains (e.g. geometry or arithmetics), language complexity (e.g. only maths, simple language, or long passages involving co-reference), question format (e.g. generative answer or fill in the blank), and background knowledge required (e.g. knowledge of formulae or commonsense). 
However, as mentioned, existing models struggle even with simple aspects that do not require background knowledge or do not contain complex language or maths. 
FERMAT complements L\={I}LA by looking in-depth at more fine-grained numerical reasoning aspects . 
It also contains expert-written templates associated with each aspect that can be used to generate an arbitrary number of new instances to address the identified shortcomings or generate more evaluation instances.
We design FERMAT for arithmetic problems presented using simple language. However, our methodology can be tailored to refine the analysis of L\={I}LA's other aspects.



\subsection{Improving Numerical Reasoning}
The literature has two main ways of improving numerical reasoning: (a) by designing task-specific models capable of numerical reasoning \cite{kumar-etal-2021-adversarial-examples, kumar-etal-2022-linguistic-augmentation, liang-etal-2022-mwp, dua-etal-2019-drop, andor-etal-2019-giving, Yang-etal-2021-NT5}, and (b) by scaling up \cite{Brown-etal-2020-GPT3, Chowdery-etal-2022-PaLM, Chen-etal-2021-codex}. Both methods also attempt to further pre-train existing models on maths related data \cite{geva-etal-2020-injecting-genbert, Cobbe-etal-2022-gsm8k, Wei-etal-2022-chain-of-thought, Lewkowycz-etal-2022-Minerva, Zhou-etal-2022-teaching-algorithmic-reasoning}. Other existing ways include using better number encoding \cite{muffo-etal-2022-calculon} or objective functions \cite{petrak-etal-2022-numerical-reasoning}.

\subsubsection{Task-specific models: Maths solvers}
Some models have been specifically created to solve maths problems by outputting expressions \cite{kumar-etal-2021-adversarial-examples, kumar-etal-2022-linguistic-augmentation, patel-etal-2021-SVAMP} or pseudo-programs \cite{liang-etal-2022-mwp, dua-etal-2019-drop} which are then evaluated using an external module. Notwithstanding the performance of these models, they can only be used to solve maths problems that, moreover, need to be represented in a closed arithmetic form. This restricts the versatility of these models both in terms of the maths and tasks that they can solve.

Unlike the other maths solvers, GenBERT \citep{geva-etal-2020-injecting-genbert} and NT5 \cite{Yang-etal-2021-NT5} generate the final output as text, making them more general-purpose. Both are pre-trained on numerical and textual tasks to solve mathematical problems.
Both of these models are evaluated on DROP \cite{dua-etal-2019-drop} which only provides an accuracy score, so their general numerical skill performance is not well-understood.\footnote{Both models report a similar performance (below 2\% difference) on DROP, therefore in our work will focus on the smaller one, NT5.}

\subsubsection{Improving maths by scaling}
More general-purpose models that perform well with respect to mathematical reasoning are GPT3 (175B) \citep{Brown-etal-2020-GPT3}, PaLM (540B) \citep{Chowdery-etal-2022-PaLM} and Codex (175B) \citep{Chen-etal-2021-codex} where their parameter size is given in brackets. GPT3 was fine-tuned by \citet{Cobbe-etal-2022-gsm8k} on GSM8K to achieve state of the art results. Similar works using PaLM and Codex investigate prompting \cite{Wei-etal-2022-chain-of-thought, Zhou-etal-2022-teaching-algorithmic-reasoning} and extended training \cite{Lewkowycz-etal-2022-Minerva}.

All of these models are general-purpose so are able to do more than solve maths problems but are not well understood. Some ablation studies analyse specific aspects of specific models. For instance, \citet{Lewkowycz-etal-2022-Minerva} conducted a digit study and highlighted that Minerva is unable to perform any multiplication of numbers with more than seven digits.
However, their sizes make it impossible for many research and industry communities to utilise them, even just at inference time. We do not have the computation resources or access for running these large models. However, FERMAT, which is publicly available and easily accessible, can be used to perform a more comprehensive analysis of these models to further identify their strengths and shortcomings. 


\section{Multi-view Evaluation Set: FERMAT}\label{Multi-view FERMAT}
FERMAT gives a holistic view of a model by evaluating fine-detailed aspects of numerical reasoning. It is akin to \citet{ribeiro-etal-2020-checklist}'s CheckList, which focuses on linguistic variations for defining its aspects.
FERMAT is used to interpret models by evaluating them on three orthogonal views including (a) Number Understanding, (b) Mathematical Operations, and (c) Training Dependency. It also provides an automated method of generating new training or evaluation examples for a given number type or operation.

We collect the initial instances for creating the FERMAT evaluation set using the established Illinois \cite{roy-roth-2016-illinois} and CommonCore \cite{roy-roth-2015-CC} datasets. After removing duplicates, we collect 1111 unique instances from these two datasets which we name the \emph{Original} set.\footnote{The \emph{Original} set acts as the comparison to existing numerical reasoning benchmarks.} We choose instances from CommonCore and Illinois because they perfectly fit with FERMAT's design by providing one or two-hop questions. Moreover, their extensive annotation is supplemented with an alignment between the numbers in the question and the corresponding expression that the solution is calculated from. We leverage these annotations in FERMAT to create different variations of the same problem for different aspects.


\subsection{Number Understanding}
Each instance of the \emph{Original} set is used to generate 18 different numerical types where the numbers change but the language is fixed. These are categorised as (a) Alternative Representations, and (b) Range of Numbers. 
Examples of each is given in Table~\ref{tab: Numerical Types}.
\begin{table}[ht]
    \centering
    \includegraphics[width=0.49\textwidth]{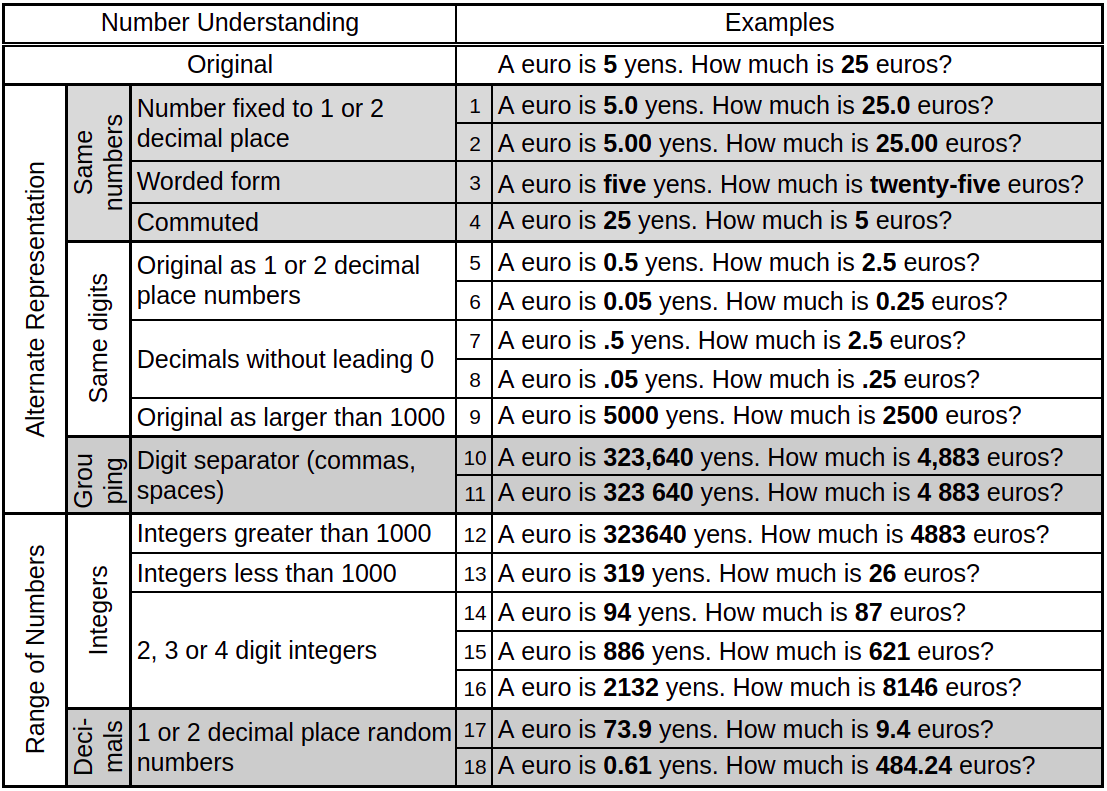}
    \caption{Numerical Types with examples.}
    \label{tab: Numerical Types}
\end{table}

\subsubsection{Alternative Representations}
Alternative Representations transforms the numbers into 11 different forms. The first four categories (rows 1 to 4) have the same number as the \emph{Original} set but represented differently whereas the next five categories (rows 5 to 9) use the same digits in the same order but by varying the magnitude of the number. The last two (rows 10 and 11) form the digit grouping subcategory where comma and space separators are used between groups of three digits.\footnote{These have different numbers to the original questions because the \emph{Original} set only contains 17 numbers where digit grouping would be visible. For comparison, the numbers are identical to the large integers type from Section~\ref{Range+of+numbers}.} This would give insight into the breadth of representations a model can accommodate, independent of the specific digit used, for instance, elucidate whether a model would be able to equally answer ``$12\times34$'', ``$34\times12$'' and ``$1.2\times3.4$''. Note that the commutative category (row 4) refers only to operations that are invariant to operand permutation and thus only has 611 associated questions instead of 1111.

\subsubsection{Range of Numbers}\label{Range+of+numbers}
The \emph{Original} set has a highly skewed distribution towards smaller integers with 94.89\% of numbers being 1 or 2 digit integers. Therefore, a random number generator is used to create 7 sub-categories of a ``Range of Numbers'' split into integers (rows 12 to 16) with large integers (greater than 1000), small integers (less than 1000) and 2, 3 and 4 digit integers, and decimals (rows 17 and 18) with 1 or 2 decimal place numbers.


\subsection{Mathematical Operations}
The operations sought by the model plays a vital role in numerical reasoning. A one-hop problem which requires a single operation, to a human, would seem much easier than a two-hop problem where an intermediate calculation would need to be computed first. With regards to this, we consider 9 operation sets generated using basic operations (addition, subtraction, multiplication and division). Their distribution is given in Appendix~\ref{app: Mathematical Operations}.

\subsection{Training Dependency Classification}\label{Training Dependency Classification}
The frequency of the occurrence of a number in pre-training data has a great impact on the performance of the model on those numbers \citep{Razeghi-etal-2022-pretraining-term-freq}. Motivated by this, FERMAT also includes a view for training dependency, but at the fine-tuning or prompting-level only. Despite the test being unseen, a model could be learning the training data and focalise on seen numbers or seen operations. Therefore, we include a Training Dependency Classification aspect to FERMAT using the following classes based on what was seen during training:\footnote{All the examples are associated to the test expression, ``$5\times(2+3)$''.}
\begin{itemize}
    \item[(a)] \emph{Exact}: all the numbers and operations are seen with the same operations modulo commutativity, e.g. ``$(3+2)\times5$'',
    \item[(b)] \emph{All Numbers}: all the numbers are seen but with different operations, e.g. ``$(5-2)\div3$'',
    \item[(c)]\emph{Number \& Operation}: at least one number and operation are seen, e.g. ``$(5+3)\div4$'', the ``5'' and the addition are at least seen, 
    \item[(d)]\emph{One Number}: at least one number is seen with none of the operations, e.g. ``$9-5$'', the ``5'' is seen but nor with the ``9'', nor with subtraction,
    \item[(e)] \emph{One Operation}: at least one operation is seen without any numbers, e.g. ``$4+7$'', the addition is seen but not with these numbers.
\end{itemize}
 It is important to note that all operations from the test set are seen in the training set, therefore according to our classification criteria, the least common class is always \emph{One Operation}. Future work may have more complicated mathematical operations in the test set that are never seen at training time such as powers or trigonometric functions, but we believe these to be too difficult for the models to learn without prior exposure. 

\subsection{Generating Training Data}\label{Automated Training Generation}
In addition to the evaluation set, FERMAT also provides a solution for generating an arbitrary length dataset that targets specific number or operation types.\footnote{In this work, it is used for training but it could also be used for evaluation.} This dataset is generated based on templates that come from three separate sources that are completely independent to the FERMAT evaluation set. The first set comprises of 100 questions written by two professional secondary school mathematics teachers and reviewed by a third one. The distribution of the templates generated reflect a uniform distribution over the operations. The second and third sources are GSM8K and AQUA where 155 and 71 templates were selected respectively. Only the questions that used at most two basic operations were extracted and the numbers were replaced by place holders to transform them into templates. These templates are only used in Section~\ref{finetune-on-augmented-training} to enhance the linguistic and mathematical variety of the templates. The distribution of operations used in the templates alongside some examples are given in Appendix~\ref{app: templates}.

\section{Experimental setup}
To demonstrate the effectiveness of our evaluation set, FERMAT, we will perform the evaluations in two settings, (a) zero-shot, where we evaluate existing models, and (b) fine-tuned, where we further train the models on arithmetic data generated using our training data in Section~\ref{Automated Training Generation}.

\subsection{Zero-shot Evaluation}
For zero-shot performance, we evaluate the following models on FERMAT without any training:\footnote{If the output of the examined model contains more than the numerical answer, e.g. the explanation of the answer, we only extract the numerical part from the generated output based on how the model is originally trained. For example, BH\={A}SKARA gives the answer before an explanation, whereas T0 provides it after.} T0~(3B) \cite{sanh-etal-2022-T0}, FLAN-XL~(3B) \cite{wei-etal-2022-FLAN}, BH\={A}SKARA~(2.7B) \cite{Mashra-etal-2022-Lila}, FLAN-large~(770M), FLAN-base~(220M), T5-base~(220M) \cite{Raffel-etal-2020-T5}, BART-base~(140M) \cite{Lewis-etal-2019-BART}, and NT5~(3M) \cite{Yang-etal-2021-NT5}, where the size of the models is given in brackets. A zero-shot evaluation is appropriate because these models are intended to be used as off-the-shelf multi-purpose models. 

T0, FLAN, BH\={A}SKARA and NT5 have been trained using prompts, so we also test them with and without prompts. We select the prompts by consulting the original papers and judge which fit closest with our question answering task (see Appendix~\ref{app: prompts} for the exact prompts used). From the models we considered, BH\={A}SKARA, FLAN and NT5 are the ones that have also been trained for maths related datasets. BH\={A}SKARA is trained on L\={I}LA and reaches near state of the art performance, thus is a reliable model to compare numerical reasoning capabilities. However, since L\={I}LA contains lots of existing data, BH\={A}SKARA has seen 46.89\% of the \emph{Original} test set \cite{Mashra-etal-2022-Lila} at training time. It also includes DeepMind Mathematics \cite{saxton-etal-2018-analysing-NN-models} in its pre-training data. FLAN has also seen DeepMind Mathematics in training. NT5 is pre-trained on synthetic numerical tasks involving non-worded problems with  integers up to 20000, decimals, negatives and percentages and textual tasks as described by \citet{geva-etal-2020-injecting-genbert}, and then fine-tuned on DROP. 

\subsection{Fine-tuned Evaluation}
For this setting, we create a training data called \emph{Base} (see Section~\ref{EXP setup: baseline}) on which we fine-tune the following models: FLAN-large, FLAN-base, T5-base , BART-base and NT5 accessed from Huggingface \cite{wolf-etal-2020-transformers-huggingface}.
We also use a digit tokeniser as implemented by \citet{petrak-etal-2022-numerical-reasoning} which gives more promising results in fine-tuning experiments compared to using the default tokeniser for numbers.\footnote{Note that NT5's tokeniser already separates the digits, so we omit the use of digit tokenisation for this model.}
Due to limitations in computational resources, we are unable to use the 3B parameter models for fine-tuning. Moreover, despite BH\={A}SKARA being advertised as a good starting point for maths related data, it is still too big for us to train.\footnote{We use NVIDIA V100 GPU nodes with a 32G memory.}


\subsubsection{Training data}\label{EXP setup: baseline}
The templates described in Section~\ref{Automated Training Generation} were used to generate the \emph{Base} training set of 200K questions with a uniform distribution over four common number types, i.e. integers and decimals with 1 or 2 decimal places all between 0 and 1000, and integers between 1000 and 1000000. This distribution also means that each of these types have 50K questions, so we would suspect that all 1000 integers between 0 to 1000 and most of the 10000 1 decimal place numbers would appear in the training set whereas all 100000 and 999900 respectively from the other two categories cannot be seen. Furthermore, all of the expert templates were used therefore the operation distribution is the same as the one for the template set (see Appendix~\ref{app: templates}). The same methodology was used to create a development set of 1K questions. This was used to decide on hyperparameters which are described in Appendix~\ref{app: Hyperparameters}. 

 
\section{Results}
\begin{table*}[ht]
    \centering
    \includegraphics[width=\textwidth]{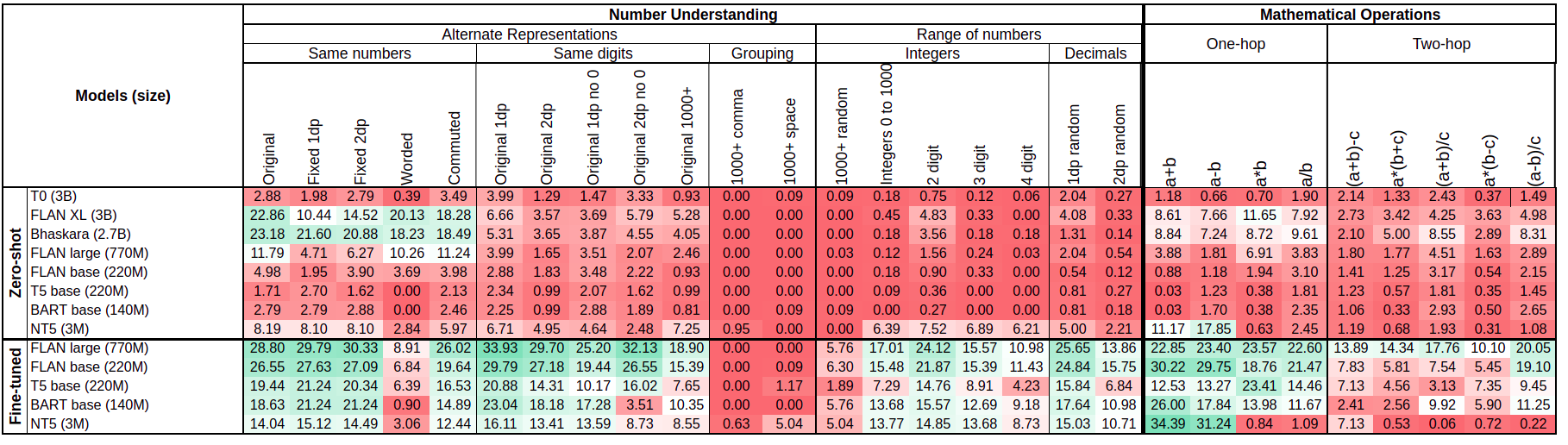}
    \caption{Zero-shot and fine-tuned performances. Accuracy shown in percentage and all green scores are above the arbitrary threshold of 10\% to subduce any false strong performances.}
    \label{tab: Zero-shot}
\end{table*}
Table~\ref{tab: Zero-shot} illustrates the zero-shot and fine-tuning performance of eight models on FERMAT with green highlighting the stronger performances for a given arithmetic type and red the poorer ones. For models that use prompts (T0, BH\={A}SKARA, FLAN and NT5), for each type, we report their mean accuracy using all the prompts and no-prompt settings. For these models, the standard deviation between the prompted and non-prompted results is below 1.5\%, therefore the reported results are representative (see Appendix~\ref{app: Zero-shot prompt} for the full results).

\subsection{Zero-shot Evaluation}
Firstly, from Table~\ref{tab: Zero-shot}'s sea of red, we can deduce that most of these models, especially T0 and the base models, tend to perform poorly at arithmetic reasoning, irrespective of size. The best-performing models, BH\={A}SKARA and FLAN-XL, are ones trained on maths data. But their performance is only respectable for a variant of the \emph{Original} set where nearly half of the numbers are single digits.

Secondly, the accuracy level for \emph{Original} is always part of the highest values, expect for NT5, so it is not a representative test set for numerical reasoning despite being derived from existing benchmarks. This could also be due to the poor diversity of the \emph{Original} set as stressed in Section~\ref{Range+of+numbers}.
Contrastingly, NT5 has its highest accuracy for addition and subtraction meaning that it is generally learning operations over specific number types.

Thirdly, even the larger models that are explicitly trained on maths datasets, i.e., BH\={A}SKARA and FLAN-XL, perform poorly on numbers that contain more than one digit indicating a limitation for their use in real-world tasks where the numbers can be of any range. This is in line with previous studies showing the shortcomings of models on longer digits \citep{Lewkowycz-etal-2022-Minerva, muffo-etal-2022-calculon}.

\subsection{Evaluation after Fine-tuning}\label{Evaluation after Fine-tuning}

As expected, with many greener cells, the fine-tuned models are better than their zero-shot counterparts and demonstrate more consistent performance across all the types. 
FERMAT's training and evaluation set templates, while covering similar aspects, are from completely independent sources. However, we observe that fine-tuning smaller commonly used models on this training data outperforms larger models like BH\={A}SKARA that are fine-tuned on various maths datasets, for instance BH\={A}SKARA is trained on over 1.32K distinct questions and programs. This underlines the benefit of creating the training data based on a diverse set of mathematical aspects. 
The larger FLAN is the only model to consistently improve on the two-hop questions suggesting that more parameters may be required to learn more complex reasoning as observed by \citet{xiong-etal-2021-multi-hop}.

Similarly, NT5 only makes significant improvement with addition and subtraction, which it was pre-trained on with synthetic questions. Therefore, as a smaller model, NT5 is only able to better generalise mathematical addition and subtraction but struggles to learn new operations during fine-tuning. However, instead of its size, this could also be due to the complexity of mathematics it has seen at pre-training.
In addition, we observe that models' performances on the ``Commuted'' aspect within the ``Same numbers'' subset are considerably lower than the other aspects. This indicates a potential for developing better number encodings that learn similar representations for the same number regardless of the position or input representation, e.g., ``three'' and 3, and 3.0. 

\subsection{Training dependency of performance}\label{training dependency of performance}
\begin{figure}[ht]
    \centering
    \includegraphics[width=0.49\textwidth]{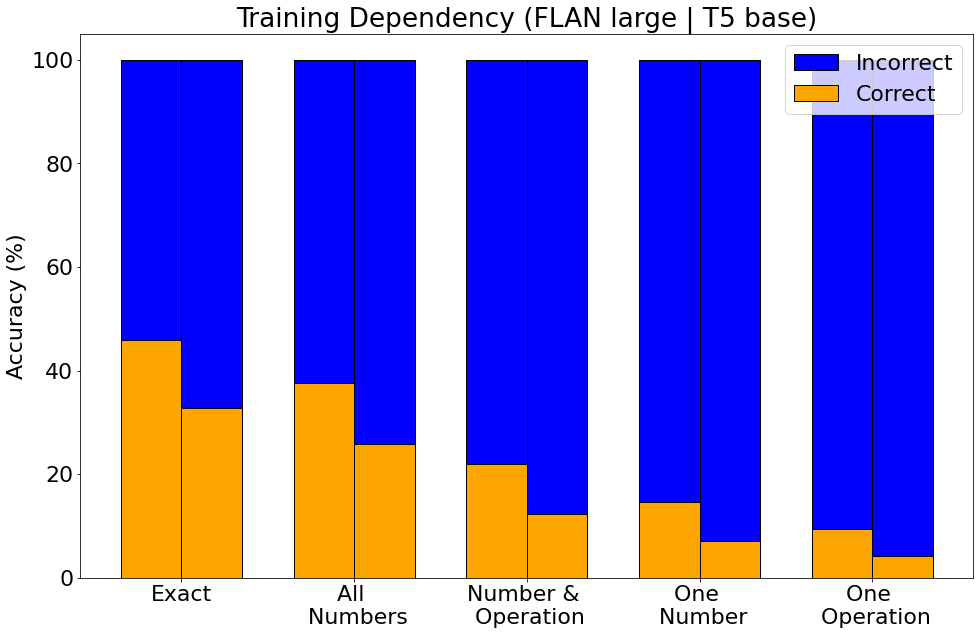}
    \caption{Training and test data overlap separated between correct and incorrect predictions made by FLAN-large (left bars) and T5-base (right bars).}
    \label{fig: Seen_unseen_bar}
\end{figure}
It is important to understand why our fine-tuned models are better across multiple types. For this, we class the expression required to answer the test sets using the Training Dependency Classification described in Section~\ref{Training Dependency Classification}. Figure~\ref{fig: Seen_unseen_bar} presents the dependency of the training data for the FLAN-large (left bars) and T5-base (right bars) models. For each bar, the ratio of correct (orange) and incorrect (blue) predicted samples are identified (the full results are given in Appendix~\ref{app: Training Dependency Results}).

The bars' monotonic trend suggests that if more of a test expression is seen at training, the model is more likely to answer it correctly. However, even for the exact match category, the performance is only 46\%. This is because the language that is used to describe the targeted equation may be different in different instances, e.g. the words ``another'' and ``increases'' are only two possible terms suggesting an addition (see Appendix~\ref{app: templates} for their use in context), indicating that the model needs exposure to a variety of different ways maths is expressed and that enriching the training data with higher language diversity can be beneficial.

In addition, the accuracy for \emph{Exact} and \emph{All Numbers} classes are similar for both models highlighting that seeing numbers during training, and therefore having a correct encoding for them, plays an important role in solving their corresponding maths operations, e.g. 89 and 30 appear both in the training set, ``\emph{Stacey prints 30 letters to post. The printer was filled with 89 sheets of paper. How many more letters could she print?}'', and in the 2 digit test set, ``\emph{89 beavers were working on their home. 30 went for a swim. How many beavers are still working on their home?}''.
This could be seconded by FLAN-large having higher accuracy than T5-base for each class as is has seen more maths at pre-training.

\subsection{Impact of training templates}\label{finetune-on-augmented-training}
As eluded in Section~\ref{training dependency of performance}, linguistic and mathematical diversity seem to be key to the improvement of numerical reasoning. Therefore, we investigate a model's performance when trained with the different templates, thus diverse language and mathematics. We fix the distribution of the aspects used in all those training instances to equal amounts of ``Integers 0 to 1000'', ``1000+ random'', ``1dp random'' and ``2dp random''. 
We use FLAN-base for the experiments of this section as it still has particularly low performances in mainly two-hop aspects according to the results of Table~\ref{tab: Zero-shot}, even after fine-tuning. Moreover, it is a small enough model to train on larger datasets.

\begin{figure}[ht]
    \centering
    \includegraphics[width=0.49\textwidth]{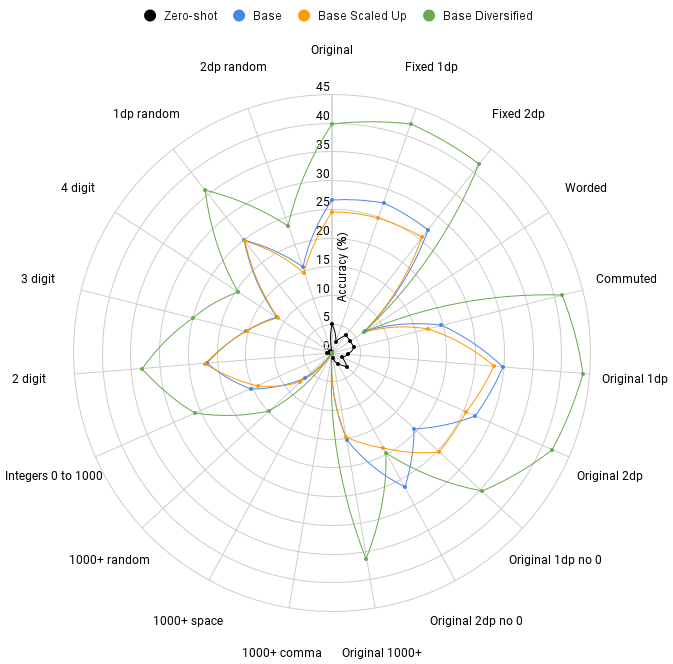}
    \caption{Fine-tuning FLAN-base on the three training sets described in Section~\ref{finetune-on-augmented-training} and the zero-shot results, see Appendix~\ref{app: FLAN-base-templates} for table of results.}
    \label{fig: Fine-tuned numerical augmentation}
\end{figure}

In this section, we consider the following three training sets to compare the effect of template diversity (see Appendix~\ref{app: training set distributions} for detailed distribution):
(1)~\emph{Base} is the 200K training data from Section~\ref{EXP setup: baseline} which only uses the expert templates,
(2)~\emph{Base Scaled Up} is \emph{Base} with an addition 100K instances from the same distribution of aspects. To make a fair comparison with the next training set, the language and mathematics is fixed as it only uses the expert templates,
(3)~\emph{Base Diversified} starts with \emph{Base} and also adds 100K instances from the same distribution of aspects. However, unlike all the other training sets which purely use the expert templates, this augments the initial set using templates recovered from GSM8K and AQUA (see Section~\ref{Automated Training Generation}) which enhances the language and mathematics seen.
We compare FLAN-base fine-tuned on the above training set along with the model's zero-shot baseline performance. Figure~\ref{fig: Fine-tuned numerical augmentation} illustrates the results of these experiments. 

First, as already established, training on diverse templates over a variety of aspects is beneficial by the shear difference illustrated by Figure~\ref{fig: Fine-tuned numerical augmentation} between \emph{Zero-shot} (black) and the fine-tuned performance (blue, orange, green). In contrast, when comparing \emph{Base} (blue) and \emph{Base Scaled Up} (orange), we remark that despite seeing 100K more combinations of numbers and operations, the learning stagnates when using the same templates meaning that the model has learnt as much as it could from the breadth of the available templates. Consequently, either linguistic or mathematical diversity is required to make a sufficient contribution. This phenomenon is, in fact, displayed by the improvement generated by \emph{Base Diversified} (green), in certain aspect by over 21\%. The diversity helps the model map the language used to describe particular mathematics better, for instance ``share'' to mean ``division'', and possibly observing more variety of this in different context seems to improve the model. Therefore, a diversity in the templates used is important, suggesting that a large variety of language may be required to attempt to further ameliorate the performance. Nevertheless, the mathematical diversity seems to also play a more important role as the diverse templates from GSM8K and AQUA have more two-hop operations (see Appendix~\ref{app: templates}). Relatedly, the mean percentage increase of one-hop operations from \emph{Base} to \emph{Base Diversified} is approximately 95\% which is about half the mean percentage increase for two-hop operations, i.e. 187\%. This suggests that mathematical variation may be more central than language diversity.

Second, the variance in accuracy between ``1dp random'' and ``2dp random'' and analogously ``Integers 0 to 1000'' and ``1000+ random'' is also intriguing. Despite having the same number of training instances with these aspects the accuracy is always lower for ``2dp random'' and ``1000+ random'' respectively, the reason for this is that these aspects involve harder skill for which either the additional 100K examples or the size of the examined model is not enough to learn this skill.\footnote{This is in line with our preliminary experiments where we observed that using complex maths datasets like GSM8K was not beneficial for general-purpose models to learn basic mathematical reasoning skills.} On the other hand, for a simpler aspect like ``2 digit'' representation, the model's performance improves considerably using the additional training instances. We can conclude that template diversity alone may not improve the models and that work on generalisation over larger sequence of integers (i.e. integers larger than 1000, more than two decimal places) such as tokenisation and representation of numbers is critical.

Third, a noteworthy observation is that \emph{Base Diversified} (green) performs worse than \emph{Base} (blue) only on the ``Original 2dp no 0'' aspect, e.g., using ``$.32$'' instead of ``$0.32$''. When further analysing the model's output of this aspect for \emph{Base Diversified}, we note that the model, on top of the 19.8\% accuracy, produces an additional 19.7\% of outputs containing correct digits but an incorrect magnitude, e.g., the correct answer might be ``$1.8$'', but the model predicts ``$0.18$''. The model might be disturbed by the decimal place or the absence of zero, implying that number encoding including positioning is vital, and thus, an accurate encoding of numbers is crucial.

\section{Conclusion}
The majority of existing datasets for numerical reasoning evaluate models based on a single score, making it impossible to identify their strengths and shortcomings to further improve them.
Multi-view benchmarks are the alternative for a more comprehensive and informative evaluation of models.
In this direction, we introduce FERMAT, a multi-view evaluation set that enables a fine-grained analysis of models based on three key aspects including number understanding, mathematical operations, and training dependency. 
FERMAT's aspects are associated with separate templates for generating instances for both evaluation and training sets, which are collected from completely independent sources and domains.

Our results confirm that comparing a single accuracy score, as with all existing maths datasets, is not representative of the performance on various numerical reasoning aspects as the evaluation dataset may be skewed towards a specific data distribution.
Based on our results, a wider language and mathematical variation can improve even smaller models. However, an apparent future direction is to focus on improving number encodings in existing models and understanding how these affect performance. 

\section{Limitations}
Three main limitations with regards to certain aspects of this paper are the comparison against very large models, the distribution of the \emph{Original} set, and the restriction of the output length.

Firstly, due to the lack of computational resources and availability of some models, we were unable to make a rigorous comparison of our fine-tuned models' as described in Section~\ref{Evaluation after Fine-tuning} against very large models like Minerva \cite{Lewkowycz-etal-2022-Minerva} or even Codex \cite{Chen-etal-2021-codex}. However, these larger models can still be evaluated as FERMAT is made publicly available.

Secondly, another limitation of FERMAT is its use of Illinois and CommonCore which have highly skewed distributions of numbers (see Section~\ref{Range+of+numbers}) and their answers are mainly integers which is not representative of the real-world. This undesired effect is mirrored in the number types that use the same numbers as \emph{Original}. However, this was part of our design for FERMAT as the alternative would have been to combined all the ranges of numbers used with the representation, creating too many aspects but mainly conflicting with non-independent analyses between representation and range of numbers. Therefore, we chose to use the same numbers as \emph{Original}, and since the templates will be openly accessible, they can be used to generate more combinations for wider aspects.

Lastly, when generating training questions, despite our best intentions, we had to limit the length of the output to an arbitrary length of 12 digits, therefore some number combination were not possible, for example $1\div 3=0.3333...$ . This practical implication could have been avoided with the use of fractions or rounding. But we judged that it would have added an extra layer of difficulty for the models and decided to restrict the output length instead.

\section*{Acknowledgements}
This work was supported by the Centre for Doctoral Training in Speech and Language Technologies (SLT) and their Applications funded by UK Research and Innovation [grant number EP/S023062/1]. Additional thanks to our mathematics teachers Ana Maria Ocampo Lucumi and Liz Scott for creating and checking the expert templates. A further acknowledgement to Constantinos Karouzos, Mugdha Pandya and Valeria Pastorino for their continued feedback in this research.

\bibliography{custom}
\bibliographystyle{acl_natbib}

\newpage
\appendix
\section*{Appendix}
\section{Distribution of Mathematical Operations}\label{app: Mathematical Operations}
Table \ref{tab:Operations_dist} gives the distribution of the various operations that exist in the \emph{Original} set and thus FERMAT's evaluation set.
\begin{table}[h!]
    \centering
    \begin{tabular}{|c|c|c|}
        \hline
        Hops & Expression & Frequency \\
        \hline
        \hline
         & $a+b$ & 154 \\
        One-hop & $a-b$ & 162 \\
         & $a\times b$ & 113 \\
         & $a\div b$ & 102 \\
         \hline
         & $(a+b)-c$ & 190 \\
         & $a\times(b+c)$& 100 \\
         Two-hop & $(a+b)\div c$ & 90 \\
         & $a\times(b-c)$ & 100 \\
         & $(a-b)\div c$ & 100 \\
         \hline
         \multicolumn{2}{|c|}{Total} & 1111 \\
         \hline
    \end{tabular}
    \caption{Distribution of the mathematical operations for the \emph{Original} set.}
    \label{tab:Operations_dist}
\end{table}

\section{Templates}\label{app: templates}
The templates' operation distribution is given by Table~\ref{app: tab: Template_distribution}. 
\begin{table}[h!]
    \centering
    \begin{tabular}{|c|c|c|c|}
        \hline
        Operations & Freq & Operations & Freq \\
        \hline
        \hline
        $\mathbf{a + b}$ & 16 & $\mathbf{a-b}$ & 28 \\
        \hline
        $\mathbf{a\times b}$ & 28 & $\mathbf{a\div b}$ & 35 \\
        \hline
        $\mathbf{a+b+c}$ & 9 & $\mathbf{a+b-c}$ & 23 \\
        \hline
        $\mathbf{a\times (b+c)}$ & 20 & $\mathbf{a\times(b-c)}$ & 13 \\
        \hline
        $\mathbf{(a+b)\div c}$ & 20 & $\mathbf{(a-b)\div c}$ & 17 \\
        \hline
        $\mathbf{a - b - c}$ & 3 & $(a\div b)+c$ & 3 \\
        \hline
        $(a\times b)+ c$ & 13 & $(a\times b)- c$ & 5 \\
        \hline
        $(a\times b)\times c$ & 10 & $(a\times b)\div c$ & 51 \\
        \hline
        $a \div (b+c)$ & 6 & $a \div (b-c)$ & 8 \\
        \hline
        $a \times (b\div c)$ & 6 & $(a\div b)\times c$ & 12 \\
        \hline
        \multicolumn{3}{|c|}{Total} & 326 \\
        \hline
    \end{tabular}
    \caption{Table of operations present in the training templates with their corresponding frequency. The ones in bold are the ones present in the expert templates.}
    \label{app: tab: Template_distribution}
\end{table}

Exemplar templates from each of three sources are given below where number place holders are in bold: \\
Expert Template: Britney has \textbf{num1} knitting needles. She buys another \textbf{num2} . How many needles does she have?\\
Expert Expression: num1 + num2 \\
\\
GSM8K Template: a trader sells \textbf{num1} meters of cloth for \$ \textbf{num2} . what is the cost price of one metre of cloth ?\\
GSM8K Expression: ( num2 / num1) \\
\\
AQUA Template: the average weight of \textbf{num1} persons increases by \textbf{num2} kg when a new person comes in place of one of them weighing \textbf{num3} kg . what might be the weight of the new person ? \\
AQUA Expression: ( num3 +( num1*num2 ))

\section{Prompts}\label{app: prompts}
Examples of the prompts used for the respective models are given below. In the examples, the underlined text is the prompt. \\
Model: T0 \\
Prompt name: Trivia\\
Example: \underline{Answer the following question. }What is 2 plus 3? \\
\\
Model: T0, FLAN \\
Prompt name: WebQA\\
Example: \underline{Question: }What is 2 plus 3? \underline{Answer: }\\
\\
Model: FLAN \\
Prompt name: Trivia \\
Example: \underline{Please answer this question: }What is 2 plus 3? \\
\\
Model: NT5 \\
Prompt name: NT5 prompt \\
Example: \underline{answer\_me: }What is 2 plus 3?

\section{Hyperparameters}\label{app: Hyperparameters}
The hyperparameters were tested on a smaller set for efficiency. During fine-tuning, we used 100 epochs with an early stopping patience of 10 and threshold of 1.0. The best model was based on accuracy of the evaluation set. All experiments were conducted with a learning rate of 5e-5, weight decay of 0.005, warm-up of 100, float32 and 3 generation beams. The rest of the hyperparameters were as the default setting in Huggingface. The max input length was 512 and max target length, 16 which is above the 12 digit limit we restrained ourselves to for the answers when generating questions. The resource used was an Nvidia Tesla V100 with 32G.

\section{Zero-shot results with and without prompts}\label{app: Zero-shot prompt}
The full results for each model including when prompts were used for all the arithmetic types are given by Table~\ref{Zero-shot-prompt}.

\section{Training Dependency Results}\label{app: Training Dependency Results}
The full results for the Training Dependency classification is shown in Table~\ref{Training Dependency table}.
\begin{table}[ht!]
    \includegraphics[width=0.49\textwidth]{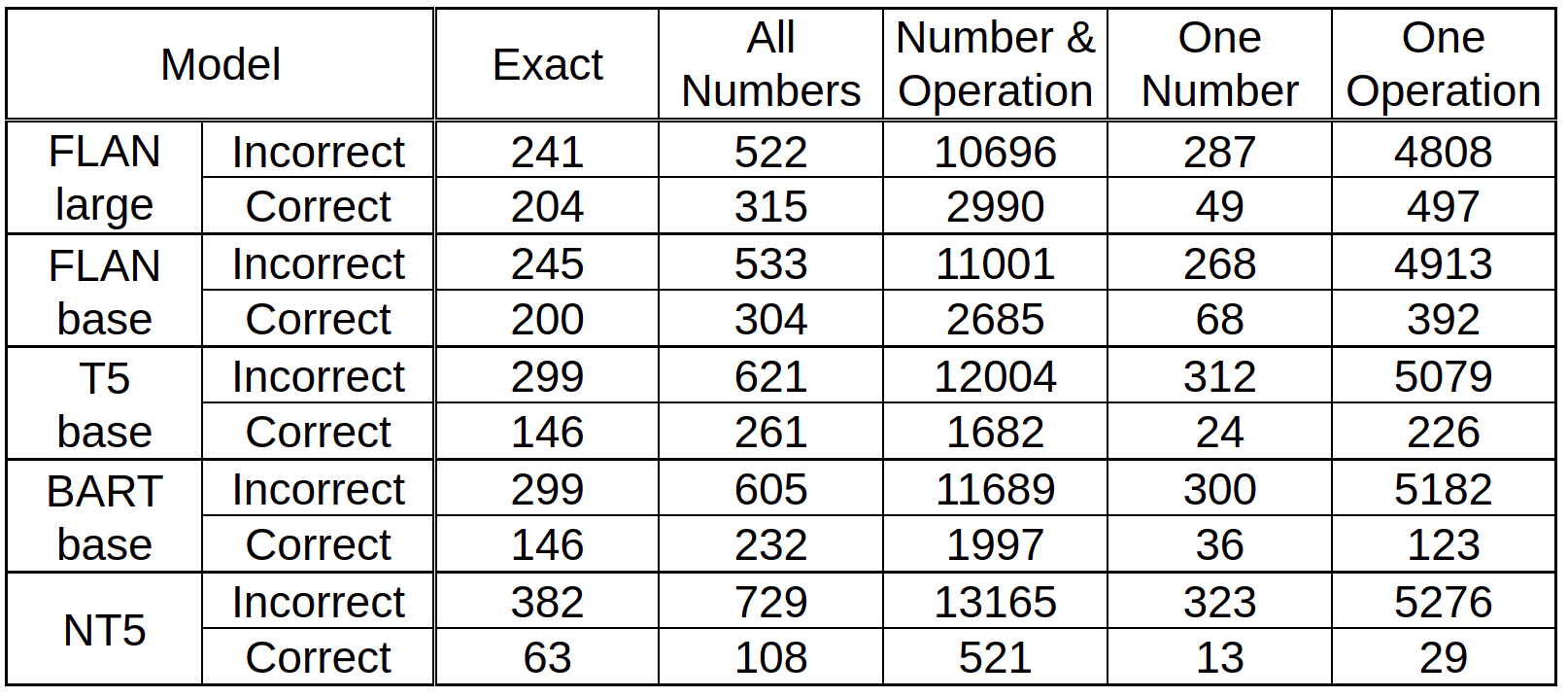}
    \caption{Training Dependency for all fine-tuned models.}
    \label{Training Dependency table}
\end{table}

\begin{table*}[ht!]
    \includegraphics[width=\textwidth]{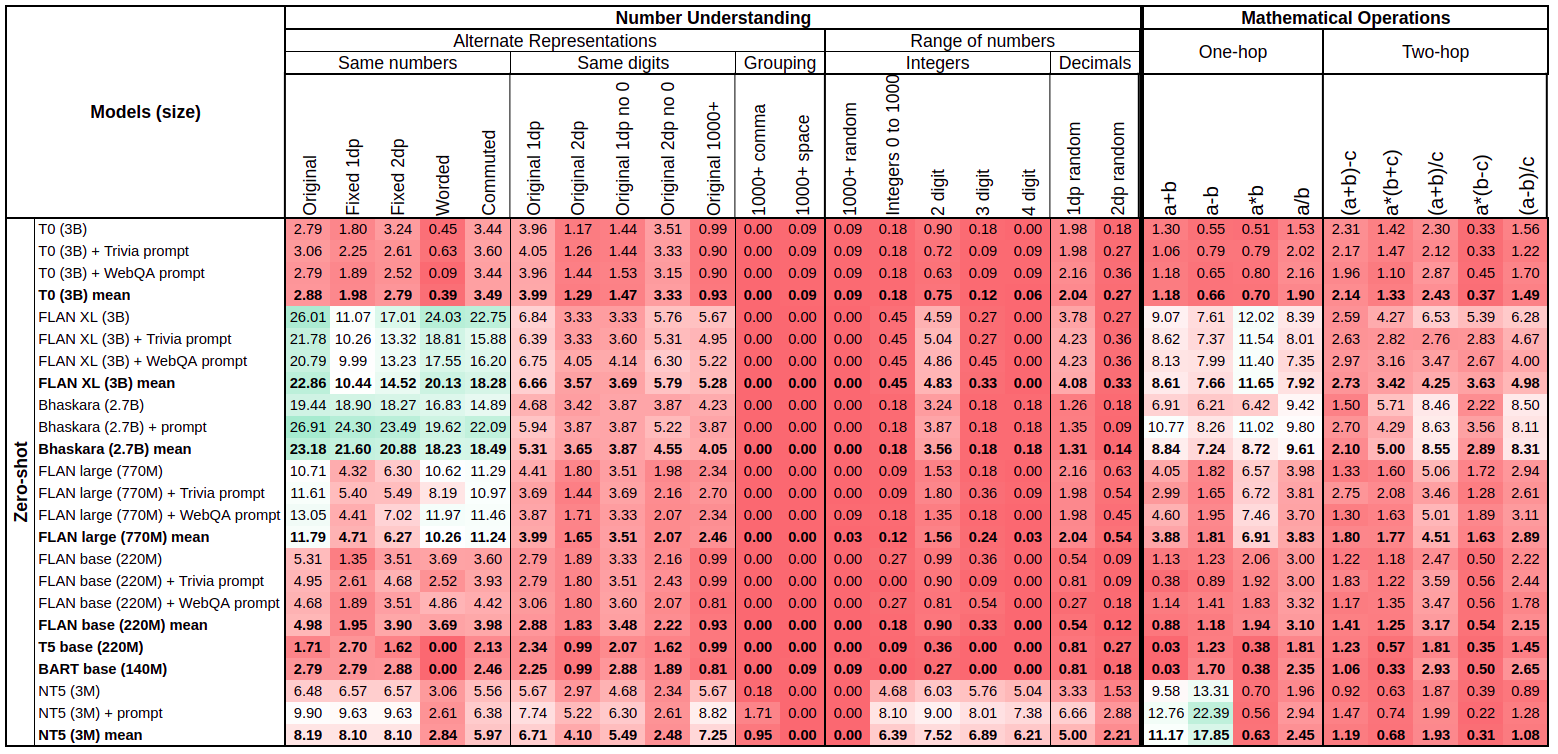}
    \caption{Zero-shot results for separate model including different prompts. Accuracy shown in percentage.}
    \label{Zero-shot-prompt}
\end{table*}

\section{Distribution of Training sets}\label{app: training set distributions}
Table~\ref{training_template} shows the distribution of the training set created from the templates, with raw numbers of instances generated based on the specific number aspect and mathematical operation design. The bold mathematical operations are the ones present in the expert templates.
\begin{table*}[ht!]
    \includegraphics[width=\textwidth]{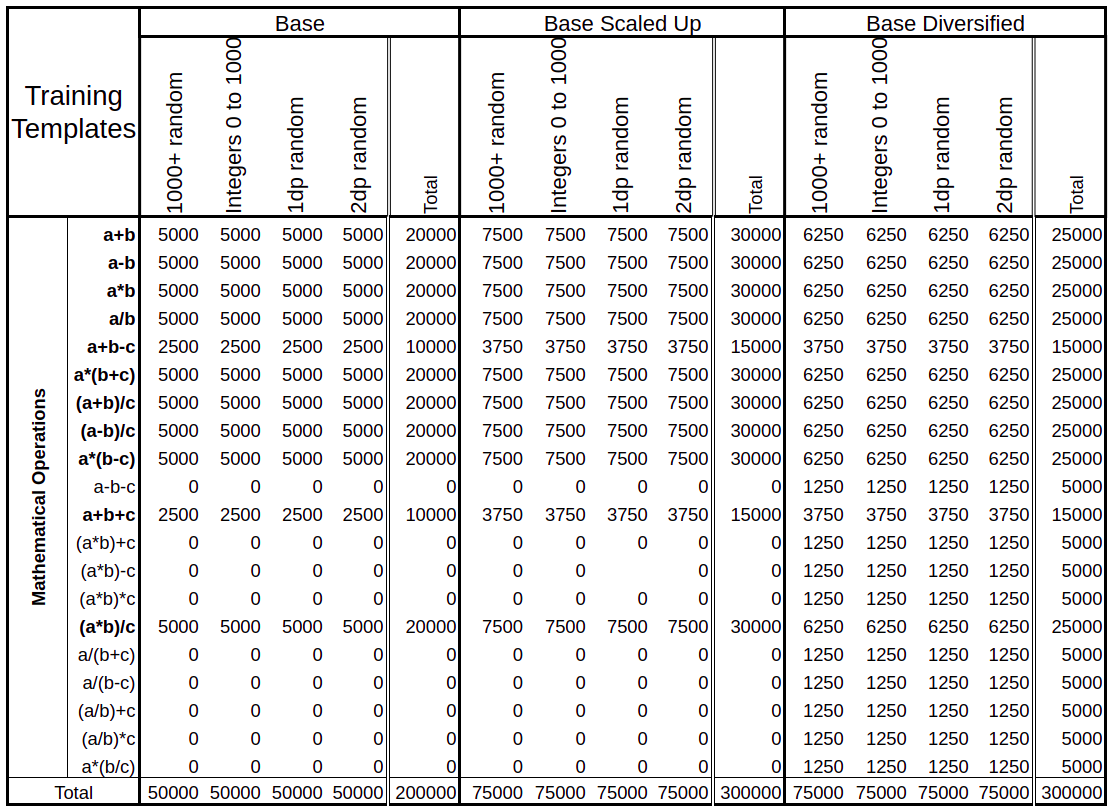}
    \caption{Distribution of templates for the \emph{Base}, \emph{Base Scaled Up} and \emph{Base Diversified} sets. In bold are the expressions that appear in the expert templates, whereas all expressions appear in the additional GSM8K and AQUA templates.}
    \label{training_template}
\end{table*}

\section{FLAN-base template diversity}\label{app: FLAN-base-templates}
Table~\ref{FLAN_base_template} shows the results of FLAN-base for each numerical reasoning aspects as a zero-shot performance and when fine-tuned on different . Accuracy is given as a percentage. Green cells indicate higher accuracy and red poorer performance.

\begin{table*}[t]
    \includegraphics[width=\textwidth]{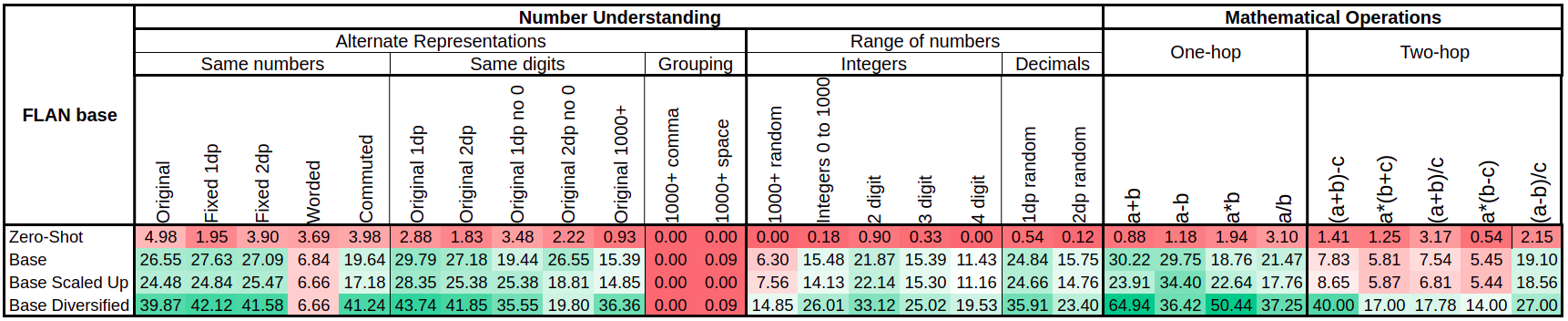}
    \caption{Results from fine-tuning FLAN-base on different distribution of templates.}
    \label{FLAN_base_template}
\end{table*}

\end{document}